\documentclass[conference]{IEEEtran}

\usepackage[cmex10]{amsmath}
\usepackage{amsthm}
\usepackage{amssymb}
\usepackage{mathrsfs}
\usepackage{graphicx}
\usepackage{float}
\usepackage{array}
\usepackage{epstopdf}
\usepackage{multirow}

\begin{document}

\title{\huge{A\MakeLowercase{utomatic} Q\MakeLowercase{uestion-}A\MakeLowercase{nswering} U\MakeLowercase{sing} A D\MakeLowercase{eep} S\MakeLowercase{imilarity} N\MakeLowercase{eural} N\MakeLowercase{etwork}}}

\author{Shervin Minaee$^*$, Zhu Liu$^{\dagger}$ \\
$^*$Electrical Engineering Department, New York University
\\ $^{\dagger}$AT\&T Research Labs\\ \\
}

\maketitle

\begin{abstract}
Automatic question-answering is a classical problem in natural language processing, which aims at designing systems that can automatically answer a question, in the same way as human does.
In this work, we propose a deep learning based model for automatic question-answering. First the questions and answers are embedded using neural probabilistic modeling. Then a deep similarity neural network is trained to find the similarity score of a pair of answer and question. Then for each question, the best answer is found as the one with the highest similarity score.  We first train this model on a large-scale public question-answering database, and then fine-tune it to transfer to the customer-care chat data. 
We have also tested our framework on a public question-answering database and achieved very good performance.
\end{abstract}

\IEEEpeerreviewmaketitle

\section{Introduction}
Question-answering (QA) is an active research area in natural language processing (NLP), which has a long history.
Baseball and Lunar were two of the early question-answering systems which answered the questions about US baseball league and analysis of rocks returned by Apollo mission respectively \cite{baseball}, \cite{lunar}.
The modern question-answering systems usually rely on a vast amount of text in a knowledge source, which can either be the information on world wide web, or some structured knowledge base such as freebase \cite{freebase}.
Many of the modern question answering systems are dealing with factoid questions. Some example factoid questions are shown below:
\begin{itemize}
  \item What currency does Germany use?
  \item When was Mozart born?
  \item Where was the first capital of the United States?
\end{itemize}

Many of the classic algorithms for question-answering consist of multiple stages, such as question processing, answer type classification and answer selection, and there is a lot of engineering involved in each step \cite{nlp}.

Many of the modern QA algorithms learn to embed both question and answer into a low-dimensional space and select the answer by finding the similarity of their features.
Due to tremendous performance of deep neural network for many problems in recent years, there have been many works using deep learning models for designing question-answering systems. 
In \cite{recursive}, Iyyer introduced a recursive neural network architecture for question answering.
In \cite{memory}, Weston proposed a memory network model for question answering, where the proposed network can reason using a long-term memory component.
In \cite{subgraph}, Bordes addressed the problem of question answering with weakly supervised embedding models.
In \cite{CNN}, a convolutional neural network (CNN) based model is proposed for modeling sentences. 
This CNN model is used by Feng \cite{ibm} for answer selection in QA systems.
Long-short term memory models (LSTM) \cite{lstm} has also been used a lot for different problems in NLP. 
Tan \cite{lstm_qa}, explored the applications of LSTM based models for answer selection.

In this work we propose a model for question-answering using a deep similarity network. First, a neural network based model is used to embed the question and answer into a low-dimensional representation.
Then, these features are fed into two parallel neural networks, and combined after a few layers of hierarchical representation to make the final decision.

The main target application of this model is for AT\&T customer  care chat system, where we use the proposed model to find the best answer from the database of old chat data, and find the confidence score of the relatedness of the selected answer for the current question. If the confidence score is larger than some threshold, it would be retrieved to the customer, otherwise the question will be referred to the human agent. 
In order to keep the users' identities and personal information protected, all the chat data used in this paper have been anonymized.
Figure 1 denotes the block-diagram of the proposed model.
\begin{figure}[h]
\begin{center}
    \includegraphics [scale=0.4] {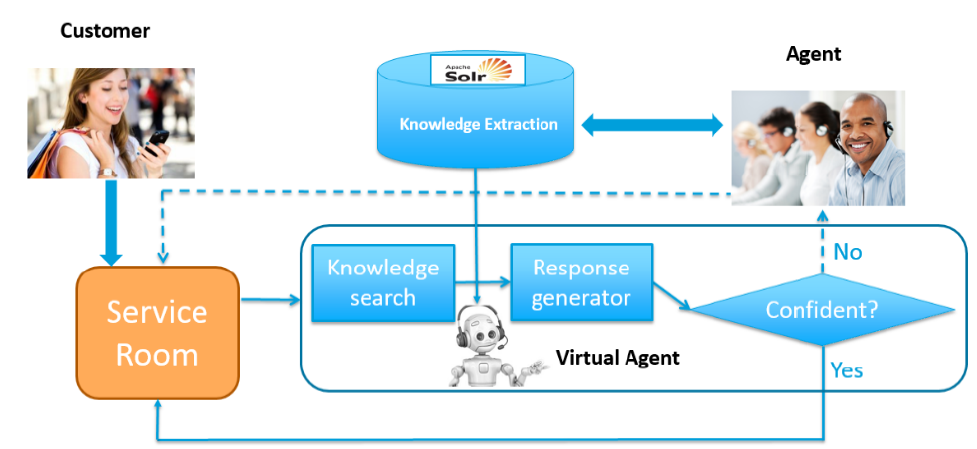}
\end{center}
  \caption{The block-diagram of the proposed model for AT\&T customer care chat system}
\end{figure}

To evaluate the performance of this model, we have tested it on a public database of insurance QA \cite{ins_qa}, and achieved promising performance on the answer selection task.

The structure of the rest of this paper is as follows: Section II presents an overview of the word and document embedding techniques. Section III describes the proposed deep similarity network architecture. Section IV provides the experimental results for the proposed algorithm. And finally the paper is concluded in Section V.

\section{Question and Answer Embedding}
To be able to process documents and text data with machine learning approaches, we need to have some vector representation for them. 
Many of the NLP tasks consider the words as the atomic units of the text data and represent each word as an index over a dictionary of vocabularies, and use bag of word technique to find the representation of each document. 

In the simplest case, where the unigram is used, each document is shown as a histogram of words appearing in it.
Despite the simplicity, this approach has some limitation: for example bag-of-words looses the order of words, and also ignores the semantic information of vocabularies. For example "soft", "food", "pasta" are equally distant from each other in bag of word representation, while the two latter ones are clearly more similar than "soft".

Because of that, there have been many works using neural network based techniques to find a word embedding that captures the similarity among words, i.e., similar words will have relatively similar representation in the embedding space. 
Here we first give a brief description of some of popular word embedding schemes using neural network, and then extend it to sentence embedding models.

Word embedding is not a new problem in NLP, and one of the early works on this area is proposed by Hinton in 1986 \cite{hinton}.
Another popular work in this area is the so called "neural network language model" (NNLM) proposed by Bengio et al \cite{bengio1}, where the word embedding is learned by a single hidden-layer neural network. 
They trained a feed-forward neural network to maximize the probability of prediction of a word, given its surrounding words.
The intuition behind this work is that given the contexts of each word (its previous and future words), there are only a small subset of words that can occur. 

In a more recent work, inspired by NNLM, Mikolov proposed a word embedding scheme, called word2vec \cite{W2V} which is more efficient than the previous ones. He proposed two models for deriving a vector representation of words, one called continuous bag of words (CBOW) and the other one Skip-gram.
In the CBOW mode, the goal is to find a representation which is good for predicting each word using its surrounding words.
In other words, for a given sequence of words $w_1$,...,$w_N$, the goal is to maximize the average log probability below:
\begin{equation}
\begin{aligned}
& {\text{maximize}}
& & \frac{1}{N} \sum_{i}^N \log p(w_i|w_{i-k},...,w_{i+k})
\end{aligned}
\end{equation}

Whereas in the  Skip-gram model the objective is to find word representations that are useful for predicting the surrounding words in a sentence or a document \cite{mikolov2}.
It maximizes an average log probability that is slightly different from CBOW, as shown below:
\begin{equation}
\begin{aligned}
& {\text{maximize}}
& & \frac{1}{N} \sum_{i}^N \sum_{j=-k, j\neq0}^k \log p(w_{i+j}|w_i)
\end{aligned}
\end{equation}

The block diagram of both of these models is shown in Figure 2. 
\begin{figure}[h]
\begin{center}
    \includegraphics [scale=0.4] {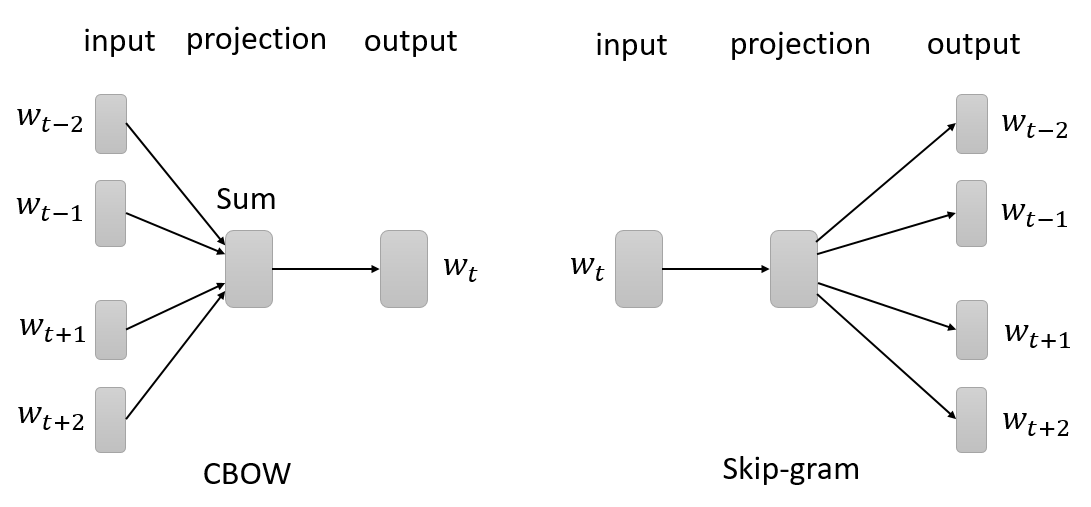}
\end{center}
  \caption{The block-diagram of the CBOW and Skip-gram models \cite{mikolov2}}
\end{figure}

These models are usually trained using hierarchical softmax regression and negative sampling technique.
After training these models, we can use the hidden layer representation as the embedding of each word, denoted by $F(.)$. 
Using this model, one can get very interesting representation of words, which can behave somehow like a vector space. 
For example if we find the word which has the closest representation to "$F(biggest)-F(big)+F(small)$", we will get $F(smallest)$.
Or $F(France)-F(Paris)+F(Germany) \approx F(Berlin)$, showing that it is also learning the semantic similarity among words.

In another work \cite{D2V}, Le and Mikolov extended this idea to find a fixed length representation for variable length texts (which could be a sentence, a paragraph, or even a document).
Similar to word2vec model, in the Doc2vec model, every word is mapped to a unique vector, showed as a column in embedding matrix $W$, and every paragraph is also mapped to another unique vector, represented by a column in matrix $D$.
These vectors are then combined (either averaged or concatenated) to predict the next word in that document \cite{D2V}.
The block-diagram of this model is shown in Figure 3.
\begin{figure}[h]
\begin{center}
    \includegraphics [scale=0.32] {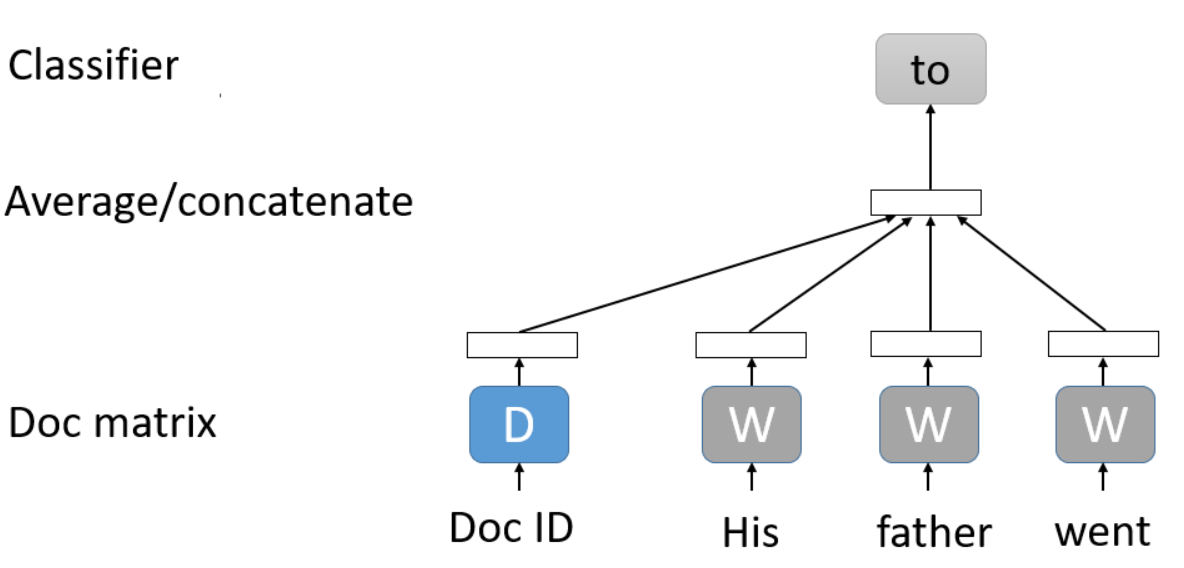}
\end{center}
  \caption{The block-diagram of Doc2vec model \cite{D2V}}
\end{figure}

After training this model, one needs to perform inference  to compute the paragraph vector for a new paragraph \cite{D2V}.
There are also many other embedding schemes using convolutional networks, recurrent nets and LSTM \cite{cnn1}-\cite{lstm_em}.

\section{Deep Similarity Network}
After extracting features we need to train a model which takes a pair of question and answer, and outputs a score that shows the properness of that answer for the given question.
There are different ways to achieve this goal. In a very simple way one could concatenate the doc2vec features of question and answer and train a classifier on top of that which predicts the probability of matching.
In this work, inspired by Siamese network by Lecun and colleagues \cite{siamese}-\cite{siamese2}, we propose a deep similarity network that takes the features for a pair of question and answer and feed them into two parallel neural networks, and combines them after a few layers of transformation to make decision.
The block diagram of this model is shown in Figure 4. 

\begin{figure}[h]
\begin{center}
    \includegraphics [scale=0.35] {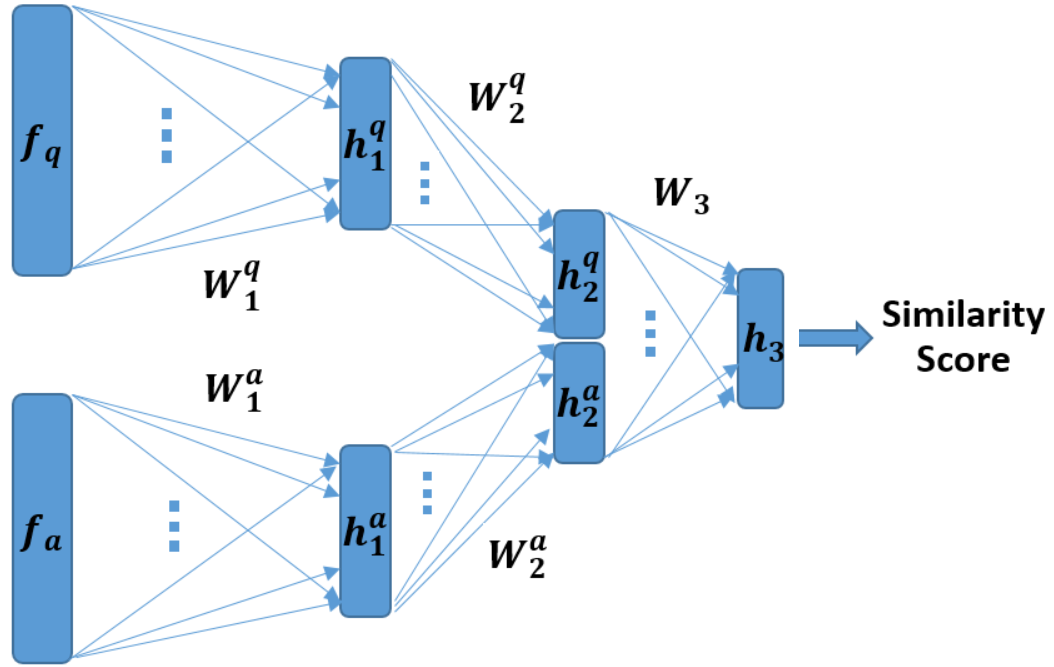}
\end{center}
  \caption{The block-diagram of the proposed similarity network}
\end{figure}

Here $f_q$ and $f_a$ denote the doc2vec features of question and answer. 
The number of layers, and the width of each layer could vary depending on the application and the size of dataset.

We would like to note that in the original Siamese network, the two networks are restricted to have shared weights, whereas in our model there is no such a restriction.
The reason is that the original Siamese network is proposed to find similarity of face (and signature in \cite{siamese2}) images, so both of its inputs are faces, and it make sense to have the same network that finds their abstract representation.
Whereas in our case one of the network is processing the question and the other one answers, and they might need different weights to map the question and answer into a common space. Also in the original Siamese network the final decision is made based on the distance between the feature embedding of  images, whereas here we propose to use the concatenated features of question and answer (i.e. $h_2^q$ and $h_2^a$ here).

This model can be trained on a training set of pairs of question and answers and a binary label showing whether this answer matches this question or not.
We train this model using cross-entropy loss regularized with the $\ell_2$ norm of weights of very last layer, as shown below:
\begin{equation}
\begin{aligned}
& \underset{W_i,b_i}{\text{minimize}}
& & \sum_{i} y_i \log(y_i^{'}) +\lambda ||W_3||_F^2
\end{aligned}
\end{equation}

\section{Experimental Results}
In this section we provide the experimental result for question classification and answer selection.
Our implementations are done in Python.
We used the Gensim package \cite{gensim} for doc2vec features extraction, and tensorflow \cite{TF} for deep similarity network training.

\subsection{Question Classification Task}
We first present a comparison between doc2vec features and bag-of-words representation for question classification, i.e., to classify whether a given sentence is asking a question or not.
We first train the doc2vec model on more than 3 million questions and answers (with a total of around 36 million words) from the AT\&T customer care chat data. 
We remove the words with very low frequency and map them to a special symbol (LF). We also map all words which have numbers into another special symbol.
We then train two doc2vec model, one for the questions and the other one for answers.

We also prepared a set of 2000 questions and answer which are manually labeled (the label indicates whether the current sentence is a question (answer) or not).
Then we use doc2vec and bag-of-word features, followed by SVM to perform question classification. 
We train the SVM model with different number of training samples. 
Figure 5 shows the classification accuracy of the doc2vec features versus bag of word for different training ratios. As we can see the doc2vec feature performs much better than bag-of-word, and its accuracy keeps increasing by using more training samples.
\begin{figure}[h]
\begin{center}
    \includegraphics [scale=0.7] {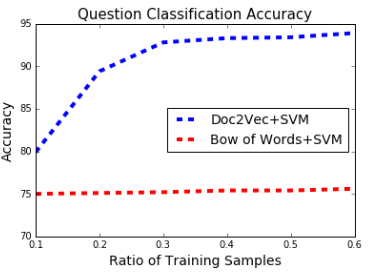}
\end{center}
  \caption{Comparison of the classification accuracy of SVM model using Doc2Vec features and Bag of Words }
\end{figure}

\subsection{Question-Answering Evaluation}
To evaluate the performance of our model for question answering, we need a database of pair of questions and answers. 
Since AT\&T chat data does not have a clean label for (question, answers) pairs, we decided to test our algorithm on InsuranceQA Corpus \cite{ins_qa}.
This database contains a training set of 12889 questions, a validation  and a test sets of 2000 questions, with pool of candidate answers for each questions. We used the data with the pool size of 100 ( on average only 65\% of questions have a correct answer in their candidates pool).
We created a set of 200k pairs of question-answer as the training set by uniformly sampling positive and negative pairs from all questions. 
We also created a validation set of 20k, and also a test set of 60k pairs of (Q,A) with a similar approach.

We used the training samples to train two doc2vec models for question and answers in this dataset.
We then trained the deep similarity network (with the first and second hidden layers having dimensions of 50 and 20 respectively) with the following hyper-parameters. 

We used a batch size of 100, and trained the network for 600 epochs with early stopping on a separate set. We used dropout with probability 0.5, and also $\ell_2$ regularization with a penalty weight of 0.0005. 
We initialized all the model weights with a zero mean Gaussian distribution with standard deviation of 0.03, and the biases with a constant value.
The learning rate is set to 0.0004, and is decayed by a factor of 0.95 after the 500-th epoch (decaying is stopped if the learning rate reaches 0.00001).
The hyper-parameters are tuned based on the performance on the validation set.

Using the proposed model, we were able to achieve an accuracy rate of 83\% on the test set.
Table I provides a comparison between the accuracy of the proposed approach and bag-of-word model for answer selection on this dataset.

\begin{table}[ht]
\centering
  \caption{Comparison of accuracy of different algorithms}
  \centering
\begin{tabular}{|m{3cm}|m{1.5cm}|}
\hline
Method  & Accuracy Rate\\
\hline 
 Bag of word+SVM &   \ \ \ \ \ \ 72\% \\
\hline
 The proposed algorithm  &  \ \ \ \ \ \ 83.4\%\\
\hline
\end{tabular}
\label{TblComp}
\end{table}

Figure 6 shows the accuracy rates on training and test sets in different epochs.
\begin{figure}[h]
\begin{center}
    \includegraphics [scale=0.28] {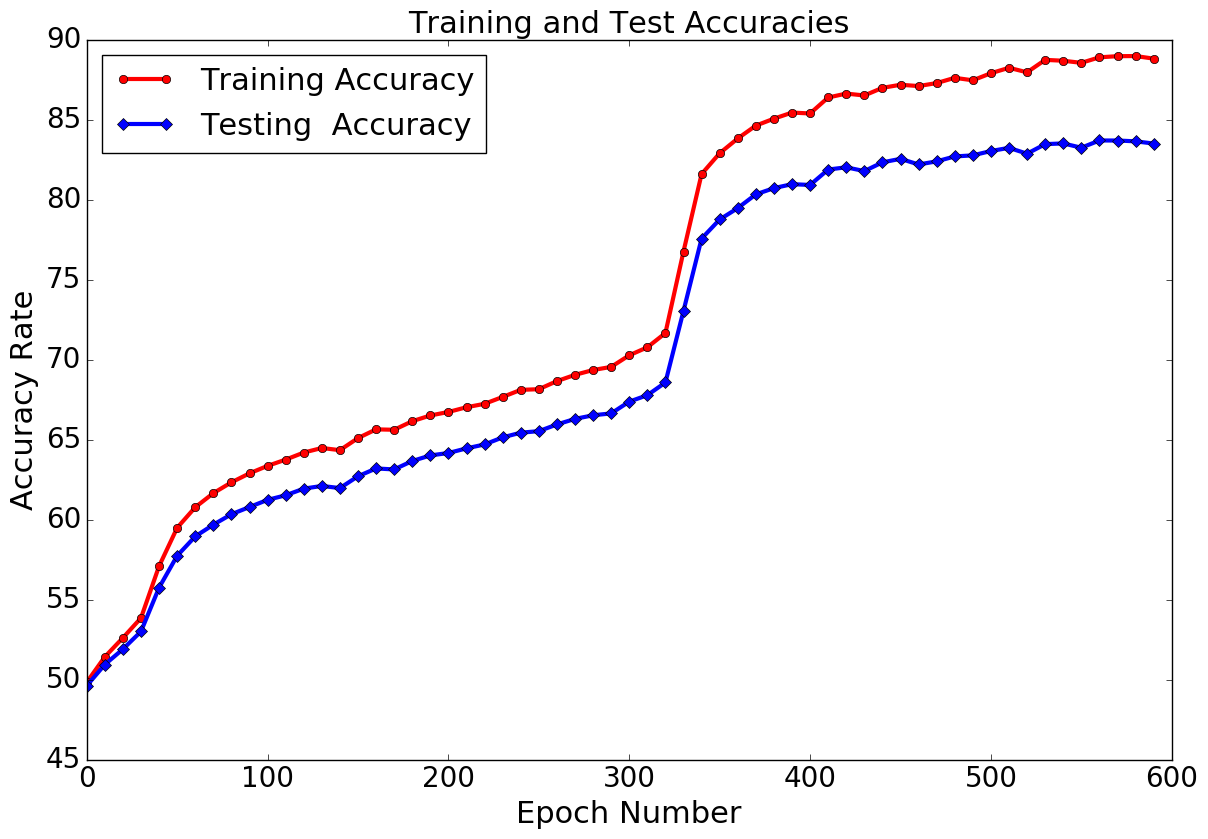}
\end{center}
  \caption{The training and test accuracies for different epochs}
\end{figure}

\section{Conclusion}
In this work, we propose a deep learning framework for question-answering, with application to customer-case service automation.
The proposed model first learns a vector representation of questions and answers using doc2vec approach.
We then train a deep similarity network, which gets a pair of question and answer as input, and predicts their similarity.
After training, this network is used to find the answer to a given question, by searching over a set of candidate answers, and retrieving the one with highest similarity score.
The proposed model is evaluated on a public database and achieved good performance.

\section*{Acknowledgment}
We would like to thank  Kyunghyun Cho at NYU for his valuable comments and suggestions regarding this work.
We also thank Minwei Feng from IBM Watson for providing InsuranceQA corpus \cite{inQA}.

\end{document}